# System and Methods for Converting Speech to SQL

Sachin Kumar[1], Ashish Kumar[2], Dr. Pinaki Mitra[3], Girish Sundaram[4]

[1]Department of Computer Science and Engineering, IIT Guwahati, Guwahati, India
`sachin.kumar@iitg.ernet.in`
[2]Department of Computer Science and Engineering, IIT Guwahati, Guwahati, India
`kr.ashish@iitg.ernet.in`
[3]Department of Computer Science and Engineering, IIT Guwahati, Guwahati, India
`pinaki@iitg.ernet.in`
[4]IBM Information Management, India Software Lab
`gisundar@in.ibm.com`

## ABSTRACT

*This paper concerns with the conversion of a Spoken English Language Query into SQL for retrieving data from RDBMS. A User submits a query as speech signal through the user interface and gets the result of the query in the text format. We have developed the acoustic and language models using which a speech utterance can be converted into English text query and thus natural language processing techniques can be applied on this English text query to generate an equivalent SQL query. For conversion of speech into English text HTK and Julius tools have been used and for conversion of English text query into SQL query we have implemented a System which uses rule based translation to translate English Language Query into SQL Query. The translation uses lexical analyzer, parser and syntax directed translation techniques like in compilers. JFLex and BYACC tools have been used to build lexical analyzer and parser respectively. System is domain independent i.e. system can run on different database as it generates lex files from the underlying database.*

## 1. INTRODUCTION

Natural Language Understanding is considered as AI Complete problem. But Queries to database are specific in nature which can be understood by the machine. Different approaches proposed for interfacing database to natural language query are mainly categorized into two techniques, one is the statistical technique and another one is classical rule based technique [4]. Statistical technique uses machine learning to process the natural language query which requires large corpus of data to train the system. Because natural language query to database are specific in nature classical rule based technique can be much faster than statistical technique. It uses the knowledge of underlying database. Our System uses the knowledge of underlying database and generate lex file automatically which will be used while tokenizing the words involved in English text query and since lex file contains underlying database information like column and table names so automatic generation of lex file helps in making the System database independent [5]. We are using six phases for conversion of speech into SQL. In first phase speech is converted into text, in second phase we analyze the text whether it is syntactically correct or not based on grammar rules for valid queries, in third phase text is mapped into an intermediate query using lexer, parser and syntax directed translation, in fourth phase we extract the SELECT clause and WHERE clause from the intermediate query, in fifth phase we find all the required tables to form the FROM clause and thus SQL query is formed, in sixth phase formulated SQL query is fired to database and result is obtained.

## 2. CHARACTERISTICS

The System converts Spoken English query into an equivalent SQL query. The main characteristics of the System are

- Provides an interface for users to query the database through either speech signal or text in English.
- The System is independent of database i.e. system can be configured automatically for different databases. It needs underlying database connection information for the configuration and underlying database schema.
- Users don't need to know the underlying database schema to query. System automatically finds the table names required to satisfy the query.
- The System uses rule based translation technique which comprises lexical analyzer, Syntax analyzer and Syntax directed translation.

## 3. COMPONENTS

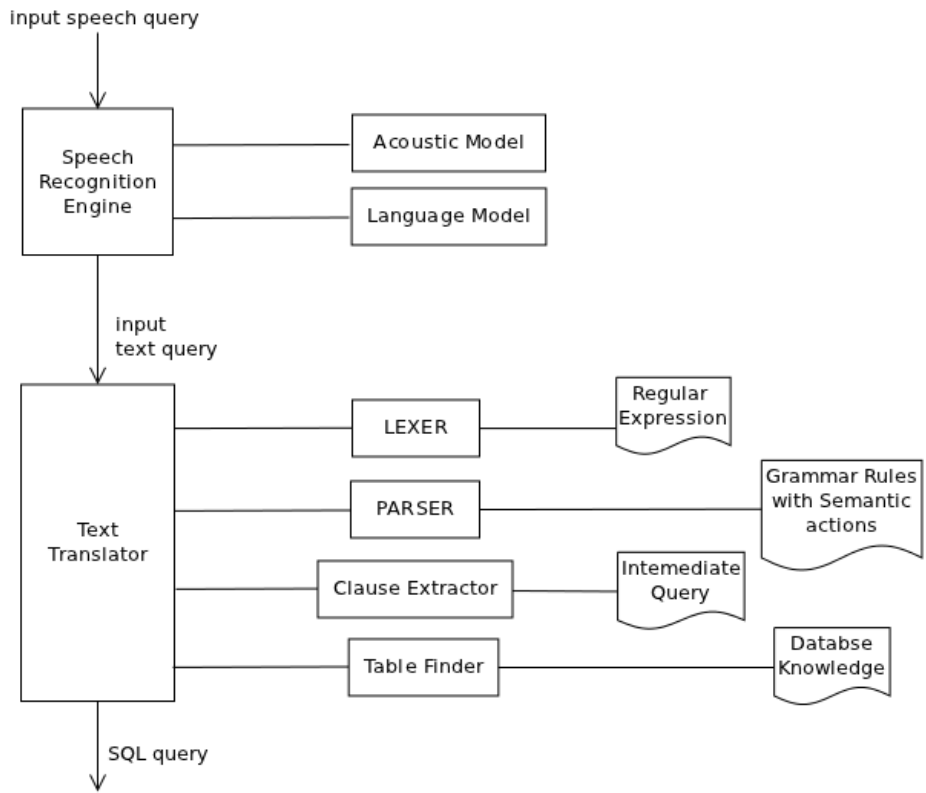

Figure 1. Components of the Proposed System

## 3.1. Speech Recognizer

Speech recognizer is used to convert a human spoken English sentence into a text string. Speech recognizer relies on acoustic and language models to convert a spoken sentence into text [1]. For this system acoustic model has been developed in HTK and the language model has been developed in Julius. Julius provides a recognizer that uses these two models to perform speech recognition [2].

### 3.1.1. Acoustic Model

Acoustic model is also known as phonetic model. Phonemes are the word units i.e. a sequence of phonemes produces a word. A phonetic model comprises a collection of states where each state represents a phoneme. Thus the phonetic model is a kind of network of states such that it can identify/accept a possible word on incoming of the input speech signal. Basically the models are developed using supervised learning during training. That is there are some training words and using the phonemes involved in each word the phonetic model is developed and at run time the speech signal of input sentence are matched against these models phoneme by phoneme to produce a correct matched word.

Since many persons can speak a same word in different quality of voices so it is not always possible to get a perfect match for the input speech signal i.e. for a spoken word if we go to identify the phoneme sequence of that word it may be possible that we could not match an exact state/phoneme in the model. Due to this, the acoustic model has been designed on the basis of Hidden Markov Model (HMM) concept. Hidden Markov model is used in a system where a state of the system is not observed perfectly i.e. there is a hidden state that we can't explore perfectly using the submitted information. A system using HMM involves hidden variables and the observed variables. The hidden variables are those that may not be perfectly observed on incoming of speech signal. The observed variables are those which have been observed on incoming of sequence of speech spectra. The observed variables are associated to hidden variables in some probabilistic manner and then generated an appropriate sequence using some probabilistic calculation. The calculation might be like "a matched hidden state is preferred for an observed state with higher probability than other matched hidden states". The core concept of HMM calculation is Baye's Rule.

A simple example of a phonetic HMM is shown in figure 2. The phonetic model is for the word "list". It can be seen in figure that after matching the phoneme 'l' at initial state we can move to the next state containing phoneme either 'ih' or 'iy' depending upon the quality and way of speaking. A phonetic HMM can be more complex including more word phonemes so there is a need of computing possible likelihood word. And this computation involves probabilistic calculation like in Viterbi Algorithm.

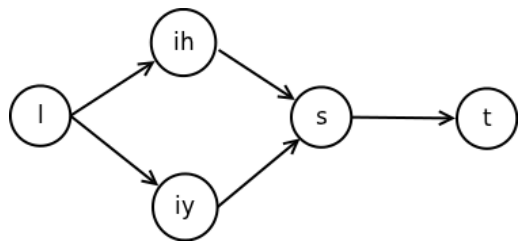

Figure 2. Phonetic HMM for word "list"

### 3.1.2. Language Model

To improve the accuracy of word recognition, a language model is used in conjunction with word HMMs. It is developed by defining grammar rules and thus specifies a valid word sequence according to grammar of the language. For a speech signal there might be more than one possible words which have been generated using phonetic/word HMMs. Then which sequence till current word is most appropriate can be chosen on the basis of syntax of the grammar that has been defined in language model.

There is also a statistical technique for developing language model known as N-gram probabilistic grammar. Examples of such models are bigram model, trigram model. In our system we have defined the language model using grammar rules. Example of grammar rules are as follows:

The processing of speech recognizer step by step can be seen in figure 3.

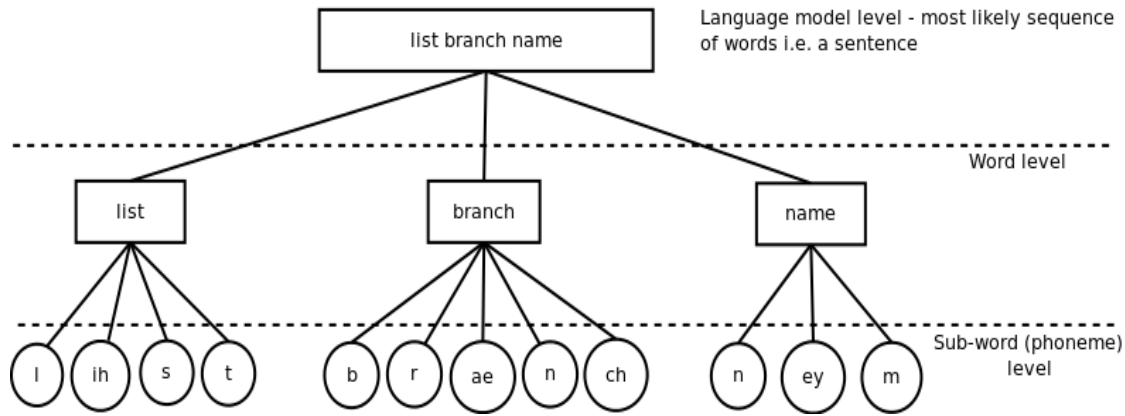

Figure 3. Formation of text sentence from Speech signal

## 3.2. Text Translator

Text translator uses classical technique i.e. grammar based technique for processing the English Text query. The translator of the system consists of lexical analyzer, parser and Table Finder for the purpose of translating the English Query into SQL query.

### 3.2.1. Lexical Analyzer

Lexical Analyzer is the first state in translation. It is called by parser to fulfil the demand of word's type. It generates the tokens for the requested words as their types and then handover to the parser for further processing. Basically JFLEX tool takes a lex file as input and generates a lexical analyzer that helps in processing of English Text query at lexical analysis phase. A lex file contains, mainly, set of regular expressions or word patterns that will be applied on each word to recognize the words as a valid token [3]. These regular expressions are written according to the need of application. JFLEX transforms the given lex file into a java class program that acts like lexical analyzer. So Lexical Analyzer performs two things, mapping of source language words to target language words and returns the appropriate token of target language word to the parser. There can be two cases while mapping:

- Map the source word to target word.
    E.g.    get     →    select
            whose   →    where

- Map the source word to nothing.
    E.g.    the     →    { }
            all     →    { }

For ex- Suppose an English query "get the branch_name"
    The possible mapping would be

    get             the             branch_name

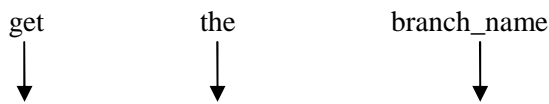

SELECT        { }                branch_name

### 3.2.2. Syntactical Analyzer

Syntax analyzer validates the English Text query whether the input query is syntactically correct. It validates the query by building a parse tree for the input English sentence. The parse tree is built with the help of the grammar rules also known as production rules. While building the parse tree Syntax analyzer also uses the concept of syntax directed translation to perform semantic analysis at the same processing phase. With the help of syntax directed translation the input English query is converted into an intermediate representation that is further processed to generate final SQL query. The parser program is generated using BYACC tool that takes a yacc file as input. The yacc file contains all grammar rules also known as production rules with associated actions [3].

### 3.2.3. Clause Extractor

The intermediate query that we got at syntax analysis phase is processed by this component and we extract the SELECT and WHERE clauses for the required SQL query from this intermediate representation. After getting these SELECT and WHERE clauses the SQL query is still not complete as FROM clause is not there in extracted query.

### 3.2.4. Table Finder

After extracting the SELECT and WHERE clauses from intermediate representation of the query we need to form the FROM clause. The FROM clause comprises all those table names which are required to fire the SQL query on the underlying database. So to find all those tables we use Table Finder component that search all columns present in the extracted query and then finds all those tables where these columns belong. After finding the tables on the basis of columns it may be possible that those tables are not directly connected in the underlying database so we need some more intermediate tables which help to connect a table to another which are not directly linked. This component finds all those required intermediate tables and returns them in the form of a path.

### 3.2.5. SQL Generator

After getting from clause if there are more than one table names then we need to perform the join operation on these tables. We are performing the join operation on the basis of common attributes between two tables. So after forming all join conditions we append these conditions in where clause.

## 4. CONVERSION STEPS

In our Experiment we have used two databases, Banking enterprise Database given in the book Database System Concept [6] and Library Database designed by us.

Suppose an example query "get customer_name whose balance is greater than 3000" for Bank database. The steps to convert this query are as follows:

1. User asks query through speech interface and corresponding text query is generated by using the acoustic and language models on which the recognizer has been trained and developed

2. Syntactic Analyzer is invoked to check whether sentence is syntactically acceptable or not and if it is acceptable then corresponding parse tree is generated. While analyzing the input sentence parser needs tokens related to words of the input sentence. For generating tokens lexical analyzer is invoked by parser. Both work in coordination until the complete parse tree is generated. The parse tree for the example query is shown in figure 4.

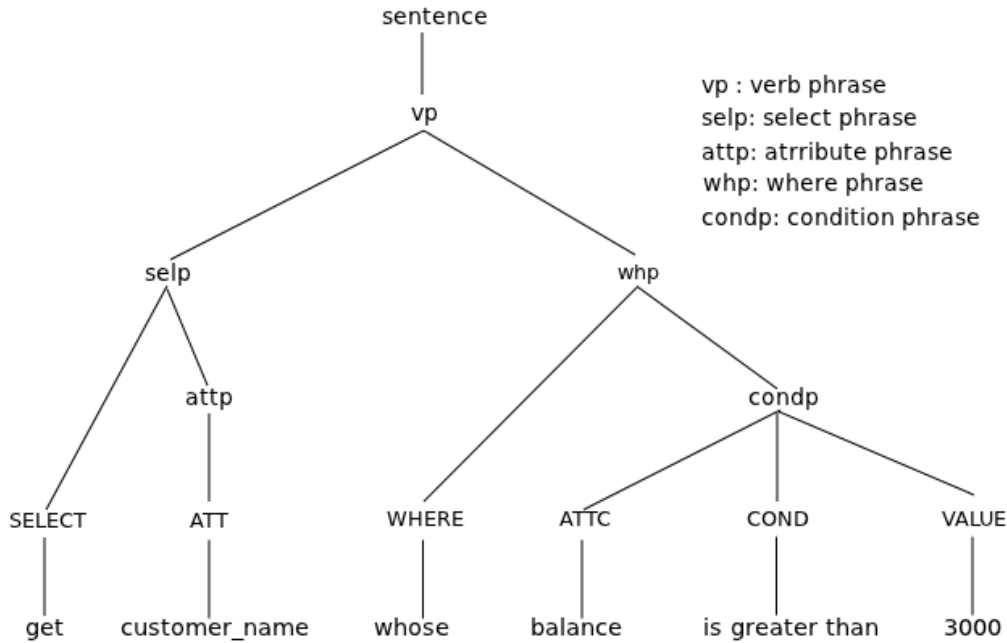

Figure 4. Parse tree of example query

3. After done with parsing we form the intermediate query representation from the parse tree. That representation is in the following form

    VP[select(customer_name), where(>(balance, 3000))]

4. We construct SELECT clause and WHERE clause from the intermediate query representation using Clause Extractor. We get the clauses for our query as follows

    selectClause: SELECT customer_name
    whereClause: WHERE balance > 3000

5. From SELECT clause and WHERE clause we identify the attributes in the query and then we find the tables containing these attributes. Attributes involved in our example query are customer_name & balance. Tables containing customer_name and balance are customer and account respectively. So we got the tables that will be involved in query.

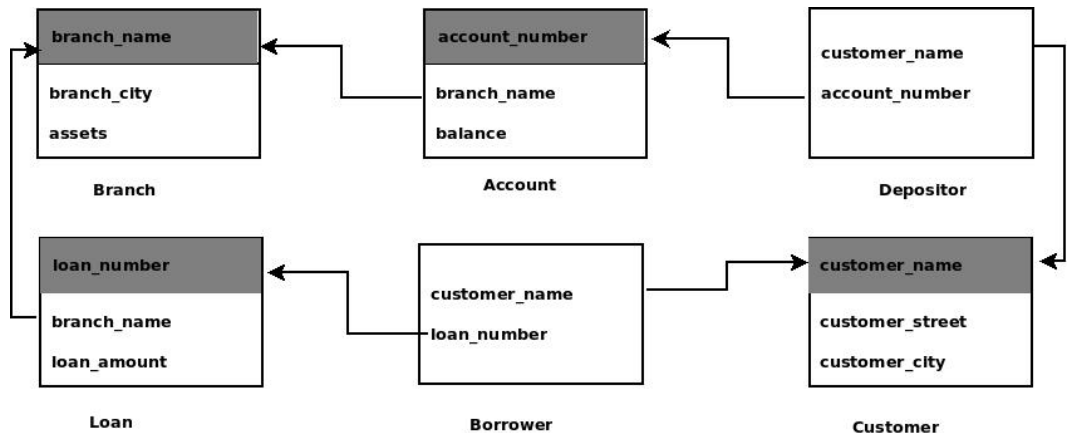

Figure 5. Schema Diagram of Bank Database

6. It may be possible that two tables are not directly connected to each other so we find intermediate tables joining these two tables. We put all these tables in from clause. Since customer and account tables are not connected directly so we need here to find the intermediate table connecting both customer and account. The intermediate table is depositor table that we can get from schema diagram using Table Finder component. The schema diagram for Bank database is shown in figure 5. So the from clause is formed with the help of these 3 tables as FROM customer, depositor, account.

7. To perform the join on two tables we need to know the common attribute of both tables to make the joining condition. Common attribute between table customer and depositor is customer_name. Common attribute between table depositor and account is account_number. So the joining conditions can be formed from these information and it is as "customer.customer_name = depositor.customer_name and depositor.account_number = account.account_number".

8. After getting the joining condition we append it to WHERE clause and our SQL query is formulated now.

Hence the formulated SQL query after the completion of these steps is

SELECT customer.customer_name
FROM customer, depositor, account
WHERE account.balance > 3000
AND customer.customer_name = depositor.customer_name
AND depositor.account_number = account.account_number

## 5. CONCLUSION

In our System we used speech recognition models in association with classical rule based technique and semantic knowledge of underlying database to translate the user speech query into SQL. User doesn't need to know the table names, we find tables with the help of attributes specified in the query. To find the join of tables we are using underlying database schema by converting it into a graph structure. System has been checked for single tables and multiple tables and it gives correct result if the input query is syntactically consistent with the Syntactic Rules.

System is also database independent i.e. it can be configured automatically for different databases.

Area in which system can be improved is the analysis and acceptance of more general query by improving the grammar.